\def\year{2021}\relax
\documentclass[letterpaper]{article} 
\usepackage{aaai21}  
\usepackage{times}  
\usepackage{helvet} 
\usepackage{courier}  
\usepackage[hyphens]{url}  
\usepackage{graphicx} 
\usepackage{subfigure}
\usepackage{amsmath}
\usepackage{natbib}  
\usepackage{caption} 
\title{Interpreting Deep Knowledge Tracing Model on EdNet Dataset}
\author{
    Deliang Wang,\textsuperscript{\rm 1}
    Yu Lu,\textsuperscript{\rm 1,2}\thanks{corresponding author}
    Qinggang Meng,\textsuperscript{\rm 2}
    Penghe Chen,\textsuperscript{\rm 2}\\
}
\affiliations{
    \textsuperscript{\rm 1}School of Educational Technology, Beijing Normal University, China\\
    \textsuperscript{\rm 2}Advanced Innovation Center for Future Education, Beijing Normal University, China\\

}

\begin{document}

\maketitle

\begin{abstract}
With more deep learning techniques being introduced into the knowledge tracing domain, the interpretability issue of the knowledge tracing models has aroused researchers' attention. Our previous study~\cite{Lu2020} on building and interpreting the KT model mainly adopts the ASSISTment dataset~\cite{feng2009addressing}, whose size is relatively small. In this work, we perform the similar tasks but on a large and newly available dataset, called EdNet~\cite{choi2020ednet}. The preliminary experiment results show the effectiveness of the interpreting techniques, while more questions and tasks are worthy to be further explored and accomplished. 

\end{abstract}

\section{Introduction}
With the fast advancements of deep learning techniques, different deep neural networks have been introduced into the knowledge tracing domain, and accordingly a number of deep learning based knowledge tracing (DLKT) models have been proposed and implemented. These DLKT models could better model the learner's knowledge state on multiple skills and capture the sequential and temporal characteristics from large scale of assessment data. Similar to other deep learning based models that work as a ``black box", the DLKT models also suffer from the interpretability issues, which has painfully impeded the deployment of DLKT models in practice. In our previous work~\cite{Lu2020}, we propose a post-hoc method to tackle the interpretability issue specifically for DLKT models, and the experiment results have shown the effectiveness of the method and its promising capabilities.

However, both the DLKT model building and the interpreting experiments are mainly based on the the ASSISTment dataset~\cite{feng2009addressing}, whose size is relatively small. In this work, we perform the model building and interpreting tasks on a new and large dataset, called EdNet~\cite{choi2020ednet}, and look into whether the similar interpreting performance can be achieved and new issues raised due to the large scale of the new dataset.

\section{Related Work}\label{sec:related-work}
Early studies adopt different machine learning models to build knowledge tracing models, where Bayesian knowledge tracing (BKT)~\cite{corbett1994knowledge} can be regarded as the pioneer work. The BKT model utilizes the hidden Markov model (HMM) to make the estimation on learner's knowledge state, and after that a number of BKT-like models were proposed and implemented~\cite{d2008more,liu2017towards,chen2017tracking,pardos2011kt,baker2011detecting}. Besides the BTK-like models, factor analysis models~\cite{pavlik2009performance} have been also proposed and well studied. The deep learning models are recently introduced into the knowledge tracing domain~\cite{piech2015deep} and subsequently, other DLKT models~\cite{zhang2017dynamic,yeung2019edm, su2018exercise, chen2018prerequisite} are proposed to keep improving the performance of learner modeling. By the end of 2020, more than 20 DLKT models have been proposed, which has played a crucial role in the knowledge tracing domain.

However, the lack of interpretability has painfully impeded the practical applications of the recent DLKT models. To tackle such a critical issue, we first propose to adopt the \textit{post-hoc} method for interpreting DLKT models~\cite{Lu2020}, where the existing layer-wise relevance propagation (LRP)~\cite{bach2015po} method has been applied to conduct the interpreting tasks. Our earlier study has shown that the proposed interpreting method is effective on the built DLKT model using the benchmark dataset ASSISTment.

ASSISTment is a commonly used dataset for knowledge tracing tasks with several different versions, namely~\textit{ASSISTment2009-2010}, \textit{ASSISTment2012-2013}, \textit{ASSISTment2015}, and \textit{ASSISTment Challenge 2017}~\cite{feng2009addressing}. Besides, there are some other well-known and public available datasets from KDD Cup2010~\cite{KDDcup} and PSLC DataShop~\cite{PSLC2010}, as well as the private datasets from Khan Academy. Among these datasets, \textit{Khan Math} dataset contains more than 47 thousand learners~\cite{piech2015deep} and \textit{ASSISTment2012-2013} has more than 2.5 million interactions after removing the duplicated records~\cite{ass12_2018}. To the best of our knowledge, the largest public available dataset in terms of learners and interactions is \textit{EdNet}~\cite{choi2020ednet}. In this work, we mainly investigate whether the proposed interpreting method is still applicable and effective on such a large dataset.

\section{Interpreting DLKT Model using EdNet}\label{sec:interprete-model}
In this section, we introduce the dataset, built DLKT model and the method to conduct the interpreting task.

\subsection{Dataset Description}
EdNet is a large-scale hierarchical dataset collecting diverse student activities from an AI tutoring service with more than 780 thousand users in Korea. It currently contains over 130 million interactions from nearly 800 thousand students over more than 2 years, which can be regarded as the largest public available dataset for building knowledge tracing model to date. In addition, EdNet records a wide variety of student actions ranging from question-solving to lecture consumption, and has a hierarchical structure dividing the student actions into four levels.

\subsection{DLKT Model Building}
We first build a DLKT model using EdNet and specifically, we employ the "KT1" dataset in EdNet. In the preprocessing step, we first filter out all the interactions whose skill tags are not available (i.e., labeled as -1), and also remove the learners who has only 10 or fewer interactions. To facilitate training and interpreting, we divide the original long interaction sequences into smaller ones with the same length of 200. As mentioned earlier, the DLKT models need to handle the sequential and temporal characteristics of learner's exercise data, and thus the RNN is often adopted, especially for the DLKT models. Hence, the built DLKT model adopts the LSTM unit, and the hidden dimensionality is set to 200. The ACC and AUC of the built DLKT model are 0.68 and 0.65 respectively. 

\subsection{LRP Method}
Similar to our previous work, we mainly adopt the LRP method to address the interpreting issue, which analyzes the contributions of input's individual features for explaining the model's decision. Briefly speaking, it firstly sets the relevance of the output layer neuron to the model's output value for the target class(es), and then starts backpropagating the relevance score from the output layer to the input layer. In different types of internal connections, the LRP would handle them in different ways, and eventually calculate the relevance values for each of the model input. These relevance values can be used to interpret the model's output directly. More technical details on interpreting the DLKT model using the LRP method can be easily found in our previous work~\cite{Lu2020}.

\begin{table}[!t]
	\centering
	\caption{Statistics of the Sequences in EdNet used for Interpreting Tasks}
	\includegraphics[width=.45\textwidth]{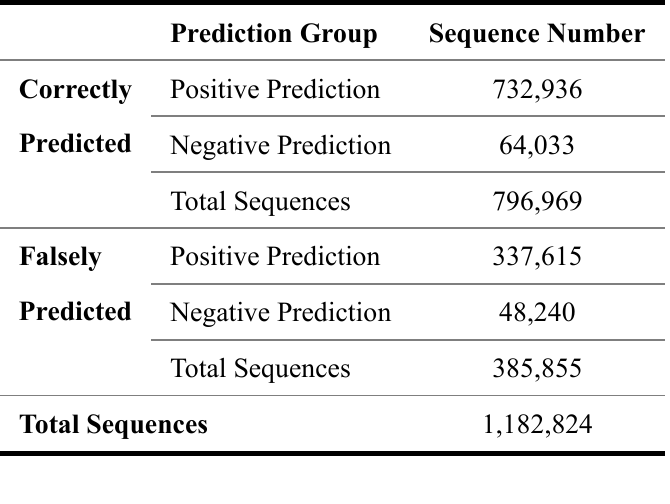}
	\label{tab:detail_infor_ednet}
\end{table}

\begin{figure*}[!t]
	\centering
	\subfigure[Positive Prediction Group]{\includegraphics [width=.49\textwidth]{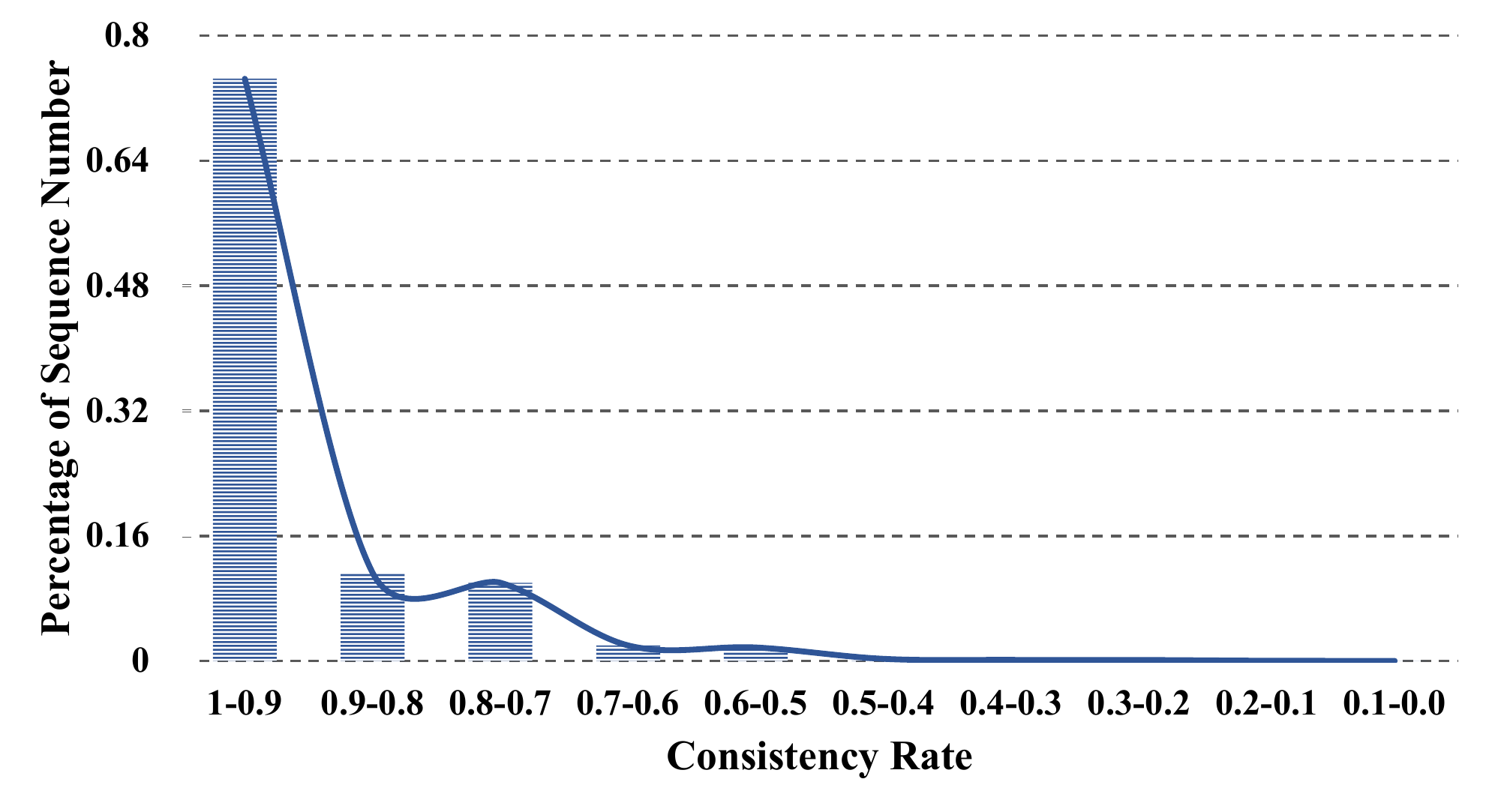} \label{cons_vs_p}}~~~~
	\subfigure[Negative Prediction Group]{\includegraphics [width=.49\textwidth]{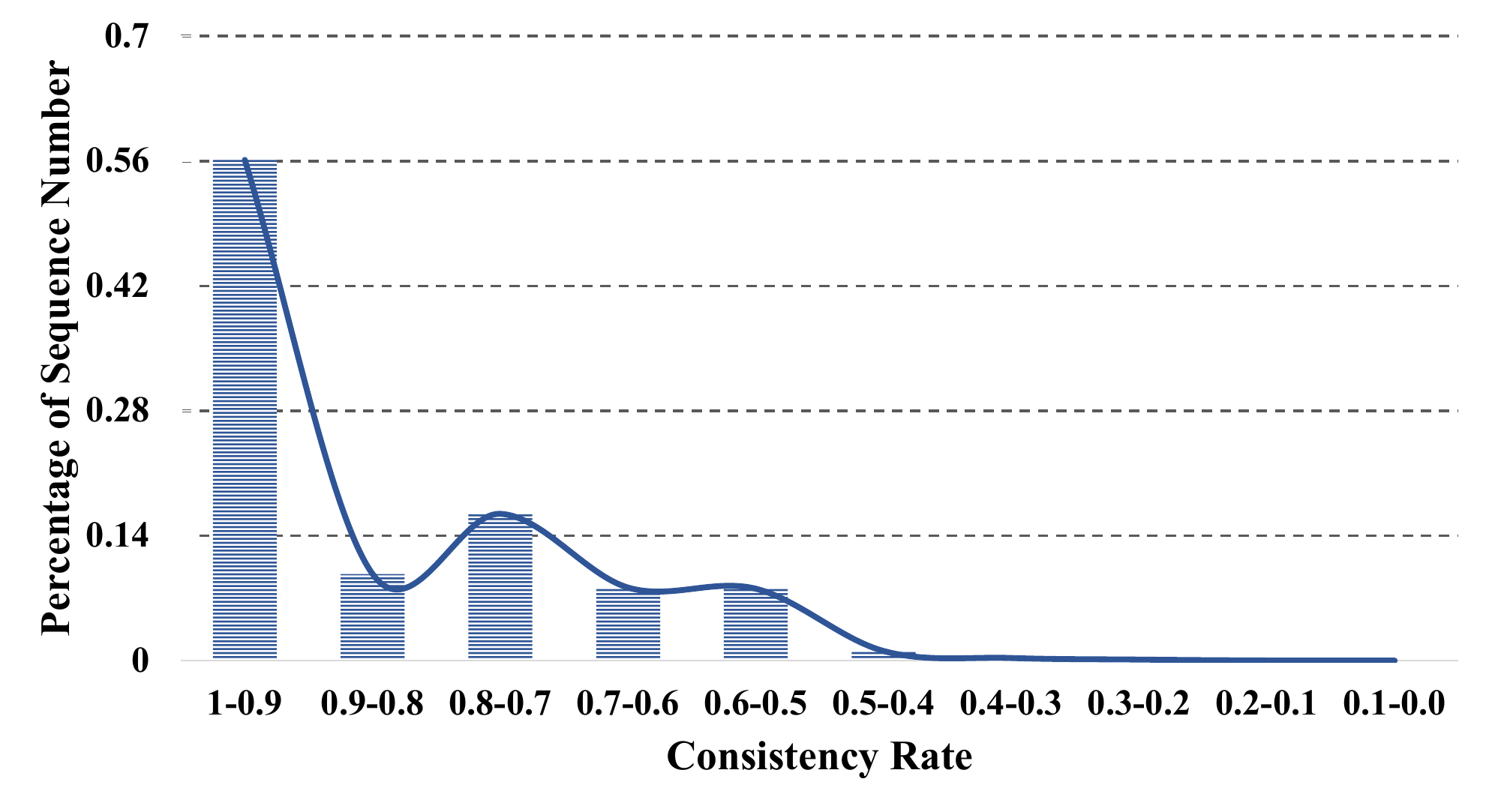} \label{cons_vs_n}}
	\caption{Histogram of the Consistent Rate on Both Positive and Negative Prediction Groups in EdNet}
	\label{fig:consistency}
\end{figure*}

\begin{figure*}[!t]
	\centering
	\subfigure[Positive Prediction Group]{\includegraphics [width=.49\textwidth]{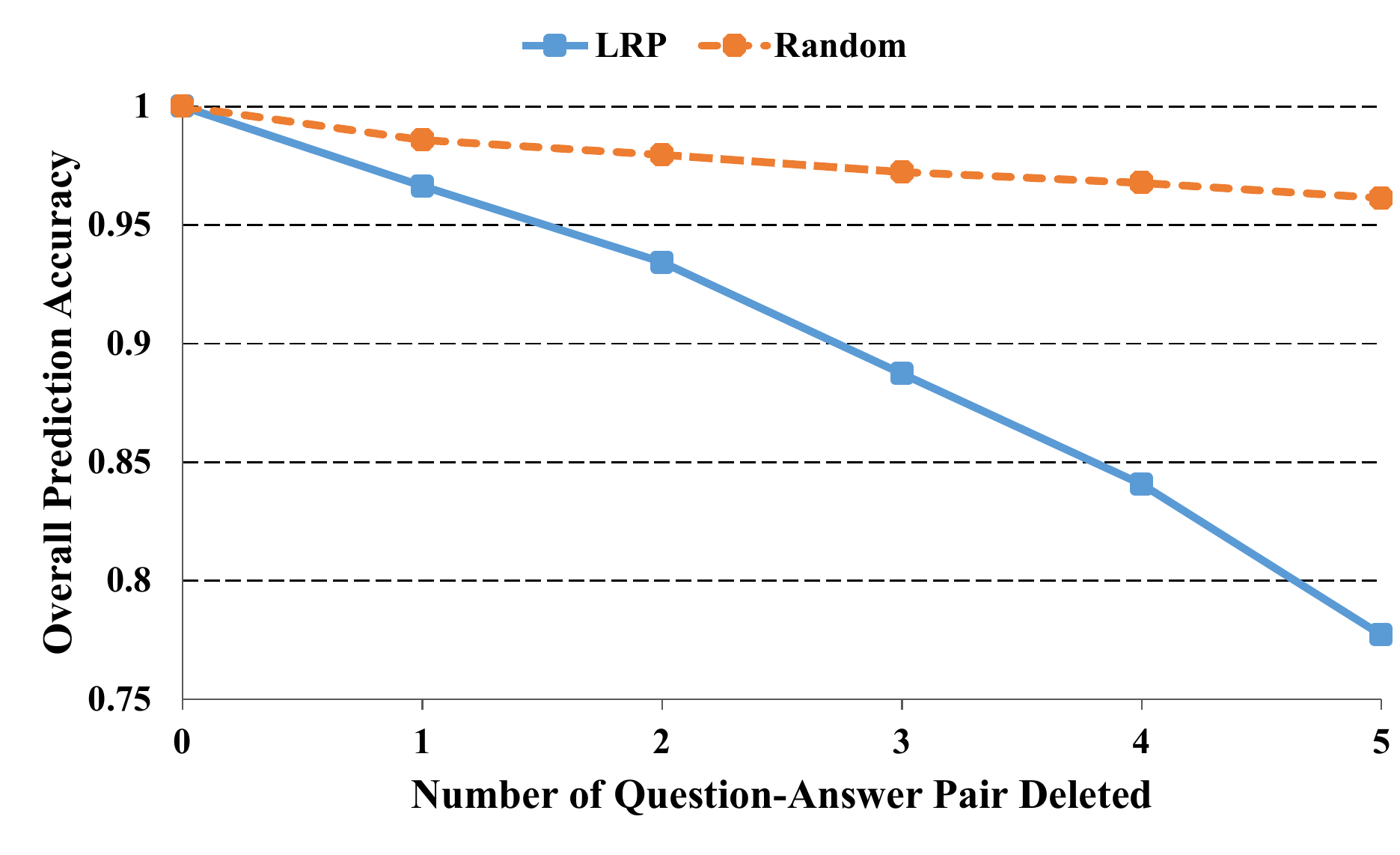} \label{vs_ass09_del_c_p}}~~~~
	\subfigure[Negative Prediction Group]{\includegraphics [width=.49\textwidth]{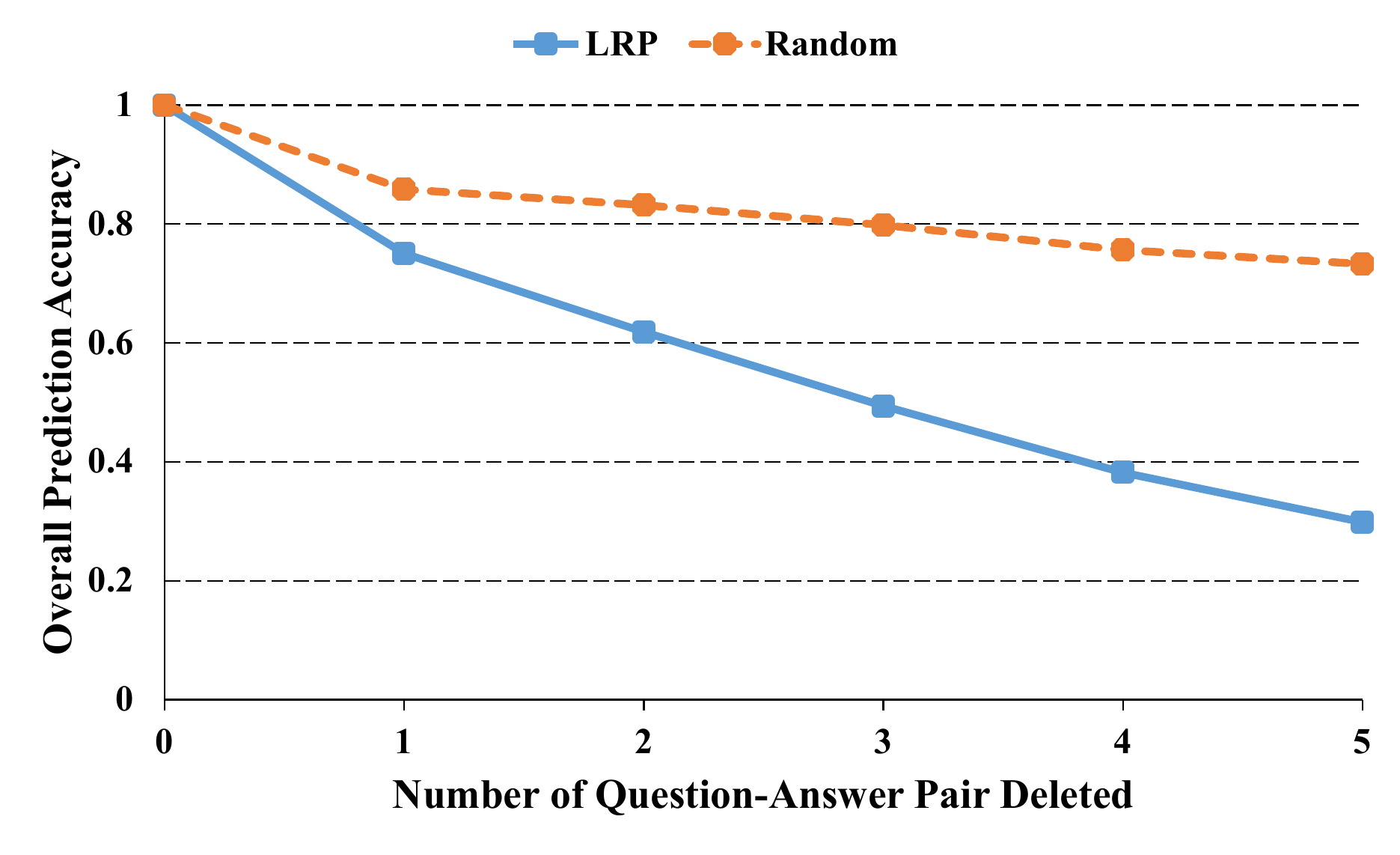} \label{vs_ass09_del_f_n}}
	\caption{Accuracy Changes on Correctly Predicted Sequences in Positive and Negative Groups}
	\label{fig:del1}
\end{figure*}

\begin{figure*}[!t]
	\centering
	\subfigure[Positive Prediction Group]{\includegraphics [width=.49\textwidth]{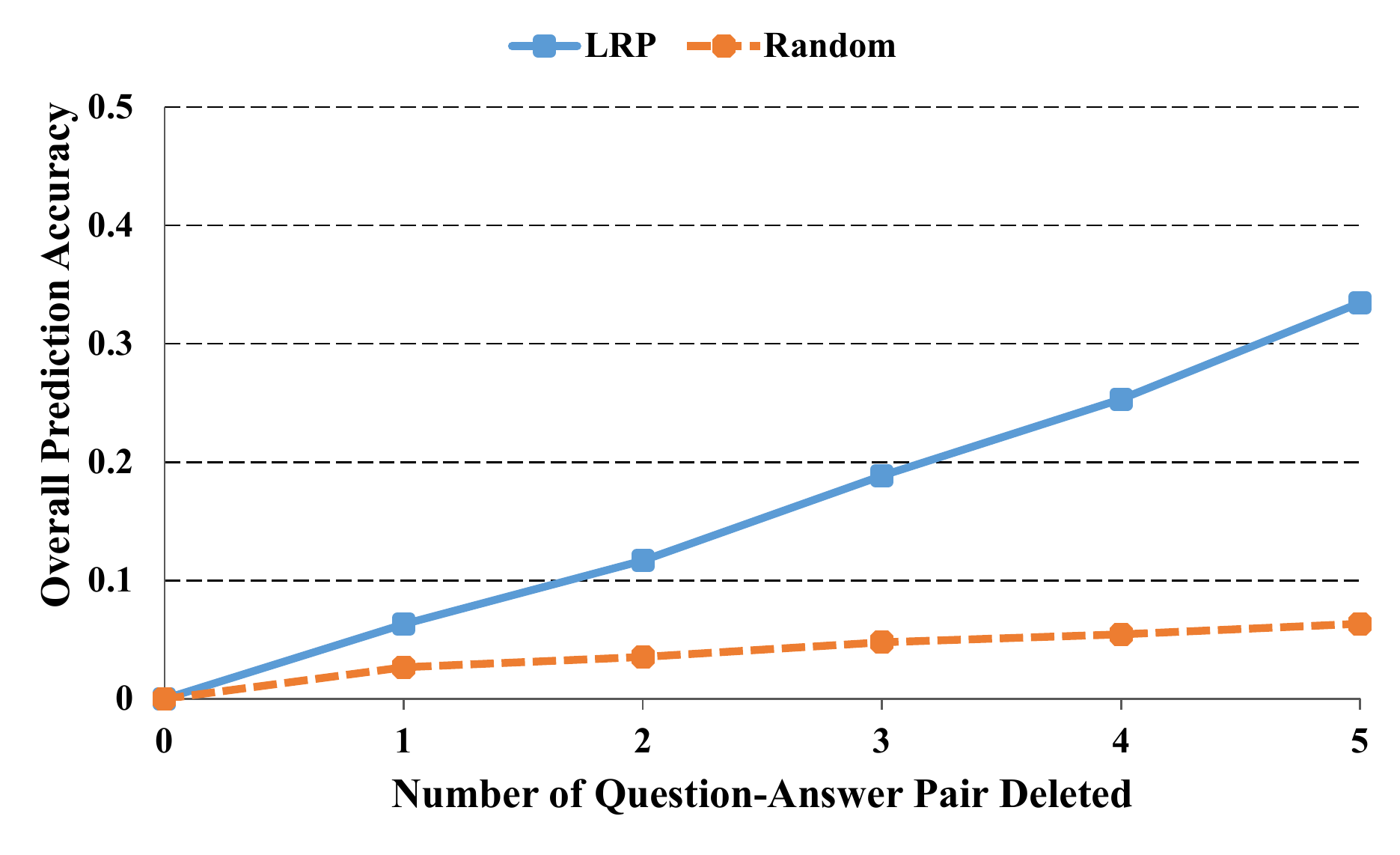} \label{vs_ass09_del_f_p}}~~~~
	\subfigure[Negative Prediction Group]{\includegraphics [width=.49\textwidth]{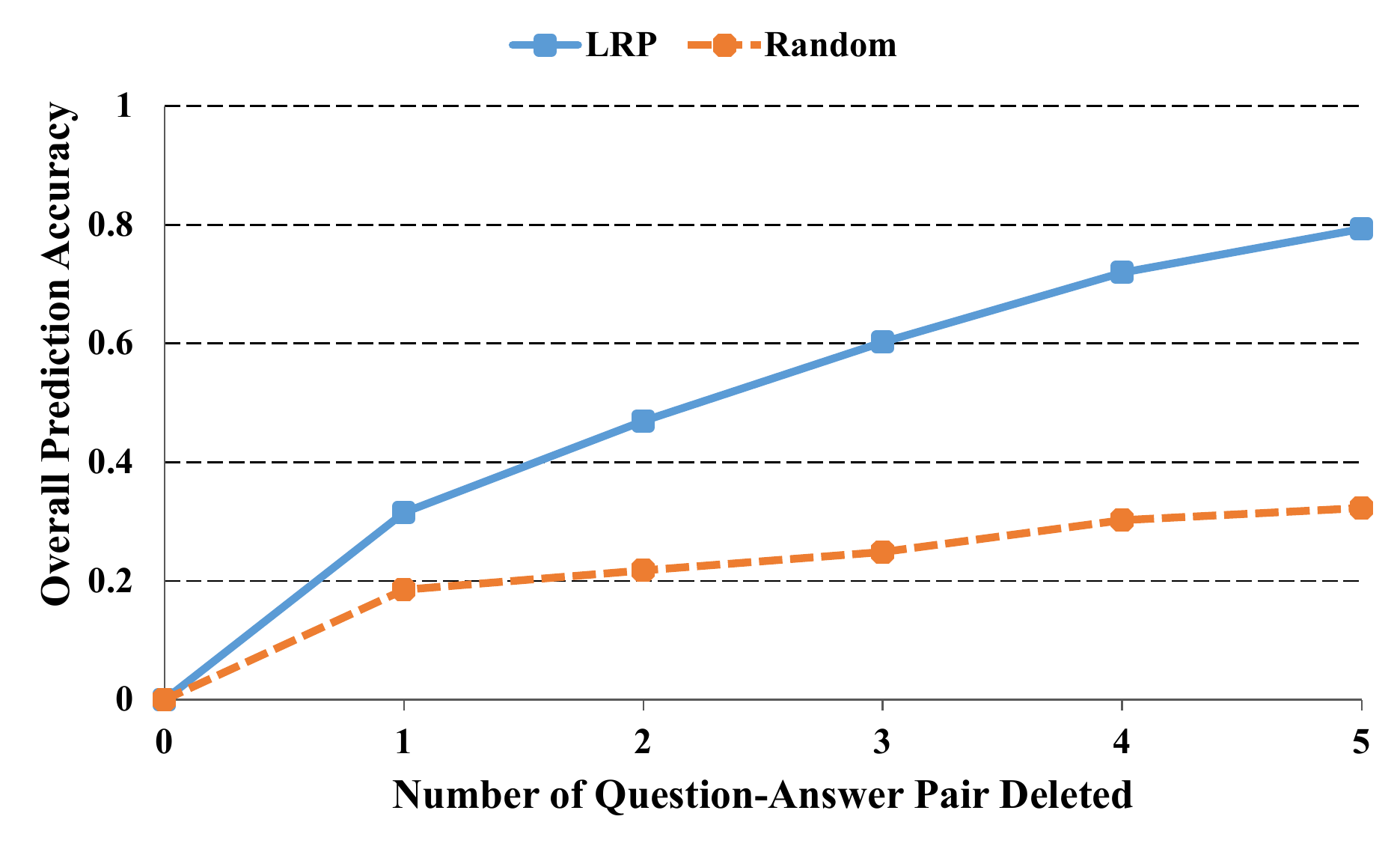} \label{vs_ass09_del_f_n}}
	\caption{Accuracy Changes on falsely Predicted Sequences in Positive and Negative Groups}
	\label{fig:del2}
\end{figure*}

\section{Evaluation}\label{sec:evaluation}
We conduct the experiments to understand the relationship between the LRP interpreting results and the model predictions. 
We divide the test data from EdNet into the sequences with a length of 15 and accordingly obtain 1,182,824 new sequences. We then take the first 14 questions of each sequence as the input to the built DLKT model, and the last one to validate the model's prediction on the 15th question. As the result, the built DKLT model correctly predicts the last question on 796,969 sequences, where the positive prediction (i.e., the mastery probability above 50\%) and the negative predictions (i.e., the mastery probability below 50\%) are 732,936 and 64,033 respectively. On the other hand, the built model falsely predicts the last question on 385,855 sequences, where the positive prediction (i.e., the mastery probability above 50\%) and the negative predictions (i.e., the mastery probability below 50\%) are 337,615 and 48,240 respectively. In the experiments, we perform the interpreting techniques to calculate the relevance values of the first 14 questions for each sequence. Using these calculated relevance values, we conduct the evaluation in two ways, namely consistency evaluation and deletion evaluation. The statistics of the dataset in term of correctly and falsely predicted sequences is summarized in Table~\ref{tab:detail_infor_ednet}.

\subsection{Consistency Experiment}
Similar to our previous work, we first investigate whether the sign of the calculated relevance values is consistent with the correctness of learner's answers. The \textit{consistent question} is defined as ``either the correctly-answered questions with a positive relevance value" or ``the falsely-answered questions with a negative relevance value". Accordingly, the \textit{consistent rate} can be defined as the percentage of such consistent questions in each sequence. Intuitively, a high \textit{consistent rate} means that the majority of correctly-answered questions have a positive contribution to the prediction result and meanwhile, the majority of falsely-answered questions have a negative contribution to the prediction result.

Figure~\ref{fig:consistency} shows the histogram of the consistent rate on both positive prediction group and negative prediction group from EdNet. Clearly, we see that around 70\% sequences in positive prediction group achieve 90 percent (or above) consistent rate, and  around 56\% sequences in negative prediction group achieve 90 percent (or above) consistent rate. Less than 5\% sequences in both positive and negative prediction groups have 50 percent (or below) consistent rate. The distribution of the consistent rate is highly similar to the interpreting results on ASSISTment dataset. It thus partially validates the effectiveness of using the proposed method to interpret the DLKT model's predictions on EdNet.

\subsection{Deletion Experiment}
Given the sign of the relevance validated by the consistency experiment above, we further investigate the quantity of the relevance (i.e., the value of the relevance) by performing the deletion experiment. In this experiment, we adopt both the correctly predicted and falsely predicted sequences as well as their relevance values calculated by the interpreting method. 

Given each of the correctly predicted sequences, we delete the questions in a decreasing order of their relevance values if it belongs to the positive prediction group, and delete the questions in an increasing order of their relevance values if the sequence belongs to the negative prediction group. Meanwhile, we perform a random question deletion for the comparison purpose. Figure~\ref{fig:del1} shows the changes in prediction accuracy of the DLKT models after deleting different number of questions on both positive and negative groups. We see that all the lines drop from the accuracy 1.0, as all the experiments are performed on the correctly predicted sequences. However, in both groups, the lines with the deletion questions using the relevance value drop much faster than the lines with the randomly deleted questions. 

Given each of the falsely predicted sequences, we delete the questions in a decreasing order of their relevance values if it belongs to the positive prediction group, and delete the questions in an increasing order of their values if the sequence belongs to the negative prediction group. Similarly, we perform a random question deletion for the comparison purpose. Figure~\ref{fig:del2} shows the changes in prediction accuracy of the DLKT models after deleting different number of questions on both positive and negative groups. We see that all the lines rise from the accuracy 0, as all the experiments are performed on the falsely predicted sequences. However, in both groups, the lines with the deletion questions using the relevance value rise much faster than the lines with the randomly deleted questions.

The above experiment results are also similar to the previous experiment results on ASSISTment dataset. Both experiments thus partially validate that the quantity of the calculated relevance reflecting the contribution to the prediction result. In other words, it is possible to use the proposed interpreting method to infer the amount of contributions of the input to the model's final prediction.

\section{Discussion}\label{sec:conclusion}
Based on our previous studies on interpreting the DLKT models, we investigate whether the same post-hoc interpreting method can be applied on a newly published large dataset EdNet. We first build a RNN-based DLKT model and accordingly conduct several experiments, including both the consistency and deletion experiments, to validate the interpreting method. We find that performing the proposed method on the EdNet dataset achieves the similar results as the ASSISTment dataset. It thus partially validates the effectiveness of the proposed method for tackling the interpretability issue of the current DLKT models.

While we have seen some promising results from the current experiments, several questions and issues are worthy to be further explored. First, the length of the sequences in EdNet is significantly large, where many of them consist of over 200 interactions. In the current experiments, we divide them into smaller ones for both model building and interpreting tasks. It is worth exploring whether the interpreting method is still effective on these long sequences. Second, the questions in EdNet are usually tagged with multiple skills, and thus how to tackle the skill-level interpretability issues on EdNet can be an interesting research problem. Finally, EdNet is organized into a hierarchical structure and each level consists of distinct types of data points. Hence, how to design a more accurate and effective interpreting method by leveraging on its hierarchical structure information might be another interesting research problem for the future work. 

\section{Acknowledgement}\label{sec:acknowledgement}
This research is partially supported by the National Natural Science Foundation of China (No. 61807003) and the Program for Student Research in the Faculty of Education of Beijing Normal University (No.1912103).

\bibliographystyle{aaai21}
\bibliography{ref}
\end{document}